\documentclass[conference,a4paper]{IEEEtran}

\usepackage[utf8]{inputenc} 
\usepackage[T1]{fontenc}
\usepackage{url}              
\usepackage{cite}             

\usepackage[cmex10]{amsmath}  
\usepackage{mleftright}       
\mleftright                   

\usepackage{graphicx}         
\usepackage{booktabs}         
\usepackage{mathtools}
\usepackage[ruled,linesnumbered]{algorithm2e}
\usepackage{dsfont}
\usepackage{amssymb}
\usepackage[dvipsnames]{xcolor}

\usepackage[suppress]{color-edits}
\addauthor{cy}{magenta}
\addauthor{yc}{orange}
\addauthor{xx}{blue}

\allowdisplaybreaks

\newtheorem{theorem}{Theorem}
\newtheorem{lemma}{Lemma}
\newtheorem{assumption}{Assumption}

\newcommand{\widesim}[2][1.5]{
  \mathrel{\overset{#2}{\scalebox{#1}[1]{$\sim$}}}
}

\begin{document}

\global\long\def\sgn{\operatorname{sgn}}%
\global\long\def\norm#1{\left\Vert #1\right\Vert }%
\global\long\def\card#1{\left|#1\right|}%
\global\long\def\cZ{\mathcal{Z}}%
\global\long\def\R{\mathbb{R}}%
\global\long\def\pmin{p_{\min}}%
\global\long\def\pthr{p_{\mathrm{thr}}}%
\global\long\def\normal{\mathcal{N}}%
\global\long\def\rank{\operatorname{rank}}%
\global\long\def\diag{\operatorname{diag}}%
\global\long\def\prob#1{\mathbb{P}\left[#1\right]}%
\global\long\def\opnorm#1{\left\Vert #1\right\Vert _{\mathrm{op}}}%
\global\long\def\ip#1{\left\langle #1\right\rangle }%
\global\long\def\nunorm#1{\left\Vert #1\right\Vert _{\ast}}%
\global\long\def\twonorm#1{\left\Vert #1\right\Vert _{2}}%
\global\long\def\fnorm#1{\left\Vert #1\right\Vert _{F}}%
\global\long\def\inftynorm#1{\left\Vert #1\right\Vert _{\infty}}%
\global\long\def\twotoinftynorm#1{\left\Vert #1\right\Vert _{2 \to \infty}}%
\global\long\def\Su{\mathbb{S}}%
\global\long\def\infnorm#1{\left\Vert #1\right\Vert _{\infty}}%
\global\long\def\argmax{\operatorname{argmax}}%
\global\long\def\argmin{\operatorname{argmin}}%
\global\long\def\ind{\mathds{1}}
\global\long\def\ME{\mathtt{MC}}
\global\long\def\cU{\mathcal{U}}
\global\long\def\cV{\mathcal{V}}
\global\long\def\diag{\operatorname{diag}}
\global\long\def\KL{\operatorname{D_{KL}}}

\title{Entry-Specific Bounds for Low-Rank Matrix Completion under Highly Non-Uniform Sampling} 


\author{%
  \IEEEauthorblockN{Xumei Xi\IEEEauthorrefmark{1},
                    Christina Lee Yu\IEEEauthorrefmark{1} 
                    and Yudong Chen\IEEEauthorrefmark{2}}
  \IEEEauthorblockA{\IEEEauthorrefmark{1}%
  School of Operations Research and Information Engineering, 
                    Cornell University, Ithaca, NY, USA, \\
                    \{xx269, cleeyu\}@cornell.edu}
  \IEEEauthorblockA{\IEEEauthorrefmark{2}%
                    Department of Computer Sciences, 
                    University of Wisconsin-Madison, Madison, WI, USA, 
                    yudong.chen@wisc.edu}
}


\maketitle


\begin{abstract}
    Low-rank matrix completion concerns the problem of estimating unobserved entries in a matrix using a sparse set of observed entries. 
    We consider the non-uniform setting where the observed entries are sampled with highly varying probabilities, potentially with different asymptotic scalings. We show that under structured sampling probabilities, it is often better and sometimes optimal to run estimation algorithms on a smaller submatrix rather than the entire matrix. 
    In particular, we prove error upper bounds customized to each entry, which match the minimax lower bounds under certain conditions. 
    Our bounds characterize the hardness of estimating each entry as a function of the localized sampling probabilities. We provide numerical experiments that confirm our theoretical findings.
\end{abstract}

\section{Introduction}
\label{sec:introduction}
    Matrix completion concerns estimating a low-rank matrix given partial and potentially noisy observations of its entries~\cite{davenport2016overview, chen2018harnessing}. This problem has applications such as in collaborative filtering~\cite{rennie2005cf}, system identification~\cite{liu2010systemid} and sensor localization~\cite{biswas2006sensor}. 
    Many algorithms with provable guarantees have been developed, including convex relaxation~\cite{candes2010matrix, candes2012exact}, alternating minimization~\cite{koren2009matrix,jain2013alternating} and spectral algorithms~\cite{keshavan2009matrix}.

   The early literature in matrix completion primarily focused on settings in which observations are uniformly distributed across the matrix, and the goal was to either derive conditions for exact recovery in the noiseless setting (e.g.~\cite{candes2010matrix, keshavan2009matrix}) or characterize the mean squared error averaged across entries under observation noise (e.g.~\cite{keshavan2010matrix}). In recent years, there has been a growing interest towards relaxing the unrealistic uniform sampling requirements as well as obtaining more fine-grained, entrywise error bounds, especially as downstream decisions may be made by comparing estimates of individual entries. While these two goals have been pursued separately, few results address both simultaneously. In particular, when sampling is non-uniform, one expects that \emph{some entries can be better estimated than the others.} Existing work falls short of capturing this phenomenon.
    
    In this paper, we tackle the above two goals jointly to answer the following research questions. Can we obtain refined \emph{entry-specific} error bounds under highly non-uniform sampling that correctly identifies the hardness of estimating each entry? Can we develop a computationally simple algorithm that is statistically efficient for estimating individual entries? When the sampling probabilities in different regions of the matrix have asymptotically different orders of magnitude, one would hope that we can retain high performance for entries in regions of the matrix with high sampling probabilities, while still providing optimal estimates for entries in regions with low sampling probabilities. Our results provide entry-specific error guarantees as a function of the localized sampling probabilities. We further show that our bounds are minimax optimal for structured sampling probabilities.

    We design a meta algorithm that can be combined with any matrix estimation method; for concreteness, we use Singular Value Thresholding (SVT)~\cite{cai2008svt}. \xxreplace{Instead of applying SVT to the entire matrix, for each target entry $(i,j)$ our method handpicks a submatrix as input to SVT, performing just as well for estimating $(i,j)$ if not better.}{For each target entry $(i,j)$, our method chooses a submatrix to input into SVT (or any matrix estimation algorithm), with the goal of obtaining a better estimate of $(i,j)$ than applying SVT to the entire matrix.}\xxcomment{Reviewer suggests rephrasing the previous sentence.} This algorithm allows us to obtain a more refined estimation error bound that has varying rates across entries. 
    We perform numerical experiments on synthetic datasets that confirm our theoretical findings.

    \subsection{Related Work}

    Several recent works consider matrix completion with the non-uniform observation pattern. Using graph limit theory, \cite{chatterjee2020deterministic} shows that a sequence of matrices is asymptotically recoverable in mean squared error if the deterministic sampling patterns converge to a graphon\footnote{A graphon is a symmetric measurable function, which serves as the limit of a sequence of dense graphs. Interested readers can consult~\cite{lovasz2012large}. } that is nonzero almost everywhere. This requirement implies the sampling cannot be too non-uniform or sparse. 
    Meanwhile, \cite{foucart2021weighted} considers non-uniform deterministic sampling patterns and proposes a simple algorithm with a weighted mean-squared error guarantee dependent on a dissimilarity function between the weights and the sampling pattern. 
    The work~\cite{cai2016matrix} studies max-norm relaxation method and shows that it achieves minimax Frobenius norm error under moderately non-uniform sampling. 
   
   A related line of work uses structured graphs to construct the sampling pattern or the weight matrix. For example, \cite{bhojanapalli2014universal} uses a bipartite graph with a large spectral gap as the sampling pattern, whereas \cite{heiman2014deterministic} uses expander graph and other graph sparsifiers.
   Complementarily, \cite{lee2013mc} considers the setting where the sampling pattern is fixed and aims to choose a weight matrix that yields weighted mean-squared error bounds.

   Going beyond (weighted) mean-squared error, several recent works consider entrywise $\ell_\infty$ error, that is, the worst-case estimation error across all entries. 
   The work~\cite{Abbe2020entrywise} provides entrywise error bounds for SVT using $\ell_\infty$ eigenspace perturbation analysis.
   Under uniform sampling, \cite{chen2020noisy} provides entrywise guarantees for convex relaxation and non-convex \xxdelete{Burer-Monteiro} approach. The paper~\cite{agarwal2021causal} considers deterministic sampling, and their algorithm searches for an almost fully observed submatrix containing the entry to be estimated. While their approach bears some similarities with ours, we note that our results consider random sampling and allow for significantly sparser observations. 

\section{Problem setup}
\label{sec:setup}

    \textit{Notation:}
    \label{subsec:notations}
        We use $c,C$ etc.\ to denote positive absolute constants, which might change from line to line. Let $[n]:=\{1,2,\dots,n\}$.
        For non-negative sequences $\{a_n\}$ and $\{b_n\}$, we write $a_n \lesssim b_n$ when $a_n \le C b_n,\forall n$, and write $a_n \asymp b_n$ or $a_n = \Theta(b_n)$ when both $a_n \lesssim b_n$ and $a_n \gtrsim b_n$ hold. 
        Let $M_{\cU, \cV}$ denote the submatrix of $M \in \R^{n\times m}$ indexed by $\cU \subseteq [n]$ and $\cV \subseteq [m]$, and $\inftynorm{M} = \max_{i , j} \card{M_{ij}}$ the entrywise $\ell_\infty$ norm. 
    
    \subsection{Latent Variable Model}
    \label{subsec:latent_variable_model}

        Our goal is to estimate the entries of a low-rank signal matrix $M^\ast \in \R^{n \times m}$ given noisy partial observations. We consider a latent variable model, where $M^\ast$ is generated via 
        \begin{equation}
            \label{eq:latent_variable}
            M^\ast_{ij} = \langle a_i^\ast, b_j^\ast \rangle,
        \end{equation}
        and the row latent variables $a_i^\ast \in \R^{r}, i\in[n]$ are sampled i.i.d.\ from some distribution; similarly for the column latent variables $b_j^\ast\in\R^{r}, j\in[m]$. If the distributions of $\{a_i\}$ and $\{b_j\}$ are sufficiently regular (e.g., sub-exponential with a non-degenerate covariance matrix), then with high probability the matrix $M^\ast$ is rank-$r$ and has a bounded incoherence parameters~\cite{vershynin2018high}. For concreteness, we consider Gaussian latent factors: $a_i^\ast \widesim{\text{i.i.d.}} N(0,I_r)$ and  $b_j ^\ast \widesim{\text{i.i.d.}} N(0,I_r)$.
        We are given a noisy and partially observed matrix
        \begin{equation}
            Y = \Omega \circ (M^\ast + E),
        \end{equation}
        where $E \in \R^{n \times m}$ is additive noise with $E_{ij}  \widesim{\text{i.i.d.}} N(0, \sigma^2)$, and $\Omega \in \{0,1\}^{n \times m}$ is the sampling/mask matrix generated as $\Omega_{ij} \sim \text{Bernoulli}(P_{ij})$, \xxedit{independently across entries}. Given $Y$, the goal is to estimate the entries of $M^\ast$. We assume $m \asymp n$ and the rank $r \ll n$ is known. 
        
    \subsection{Monotone Sampling Probabilities}
    \label{subsec:monotone_sampling_pattern}

    In the above model, the observations are non-uniform as determined by the sampling probability matrix $P=(P_{ij}) \in [0,1]^{n \times m}$. We assume that $P$ is known, which is a reasonable assumption in settings where the learner has (partial) control over the sampling process or can estimate $P$ from data.
    \xxreplace{Note that  sampling with an arbitrary $P$ is the same as an arbitrary deterministic sampling pattern $\Omega$}{Without restriction on $P$, this setting includes arbitrary deterministic sampling pattern as a special case}\xxcomment{rephrase ``the same''}---just let $P_{ij}$ be binary---under which matrix completion is NP-hard~\cite{hardt2014computational}. Therefore, we further assume $P$ has a monotone structure: there exist  permutations $\pi_n: [n] \to [n]$ and $\pi_m: [m] \to [m]$ such that $P_{\pi_n(i) \pi_m(j)} \ge P_{\pi_n(i') \pi_m(j')}$ whenever $\pi_n(i) \le \pi_n(i')$ and $ \pi_m(j) \le \pi_m(j')$. Without loss of generality, we may assume both $\pi_n$ and $\pi_m$ are the identity:\footnote{If $\pi_n$ and $\pi_m$ exist, they can be found by sorting the rows of $P$ and then the columns.}\xxcomment{Reviewer: Can anything be said about the computational complexity of finding $\pi$ in the general case?}
    \begin{assumption}[Monotonicity]
            The probability matrix $P$ satisfies $P_{ij} \ge P_{i'j'}$ if $i \le i'$ and $j \le j'$. 
            \label{assum:mono}
    \end{assumption}
        
    Assumption~\ref{assum:mono} is satisfied, e.g., in a movie rating setting, where the probability of user $i$ rating movie $j$ is determined by the activeness of the user and the popularity of the movie. In fact, this example corresponds to a special case of Assumption~\ref{assum:mono} where $P$ has rank one, as stated below. We sometimes consider this  stronger assumption.
        \begin{assumption}[Rank-one $P$]
           There exist vectors $\alpha=(\alpha_1, \dots, \alpha_n)$ and $\beta=(\beta_1, \dots, \beta_m)$ such that
            \[
                P_{ij} = \alpha_i \beta_j, \quad \forall (i,j) \in [n] \times [m],
            \]
            where
            $ 1 \ge \alpha_1 \ge \cdots \ge \alpha_n \ge 0$ and $1 \ge \beta_1 \ge\cdots \ge \beta_m \ge 0$.
            \label{assum:rank1}
        \end{assumption}

\section{Algorithm: submatrix completion}
\label{sec:algorithm}
    Let $\ME(\cdot)$ be a black-box matrix completion subroutine, e.g., SVT. If $\ME(\cdot)$ were applied on the entire observed matrix $Y$, it outputs an estimate $\hat M = \ME(Y) \in \R^{n\times m}$ of the true signal matrix. Our algorithm, \emph{submatrix completion}, instead applies $\ME(\cdot)$ to carefully chosen submatrices of $Y$.
    In particular, for each target entry $(i,j)$ to be estimated, we compute an index
    \begin{equation}
        \label{eq:kast_formula}
        k^\ast \equiv k^\ast(i,j) = \underset{k \le \min \{n,m\}}{\argmax} \ k \cdot \min \{P_{\max \{i,k\}, k}, P_{k, \max \{j,k\}}\}.
    \end{equation}
    We then apply $\ME(\cdot)$ on the submatrix indexed by $[k^\ast ] \cup \{i\}$ and $[k^\ast] \cup \{j \}$, and use the corresponding entry of the output $\ME(Y_{[k^\ast] \cup \{i\}, [k^\ast] \cup \{j \}})$ as an estimate of the target entry $M^\ast_{ij}$. In the optimization problem~(\ref{eq:kast_formula}), the variable $k$ corresponds to the size of the submatrix used to estimate $(i,j)$, $P_{\max \{i,k\}, k}$ is the smallest probability on the last row of the submatrix excluding entry $(i,j)$, and $P_{k, \max \{j,k\}}$ is the smallest probability on the last column. As will become clear in Section~\ref{subsec:upper_bound}, $k^\ast$ is chosen to minimize an upper bound on the entrywise estimation error of the submatrix.

   For illustration and ease of analysis, we adopt SVT as the matrix completion subroutine $\ME(\cdot)$. 
   Given the observation $Y$, SVT forms the rescaled observation matrix $\bar Y = (Y_{ij} / P_{ij})_{i\in[n],j\in[m]}$ (which is an unbiased estimator of $M^\ast$), and then computes the best rank-$r$ approximation $\hat{M}$ of $\bar Y$.
    Explicitly, if $\bar Y$ has singular value decomposition (SVD) $\bar Y = \bar U \bar \Sigma \bar V^\top$, where $\bar\Sigma$ is a diagonal matrix containing the singular values of $\bar Y$ in descending order, then $\hat M = \bar U_{\cdot [r]} \bar \Sigma_{[r], [r]} \bar V_{\cdot [r]}^\top$.

    \subsection{Illustrating example}
    \label{sec:example}
    
     We illustrate how our algorithm works with a concrete example. To this end, let us first derive a useful characterization of the solution $k^\ast(i,j)$ to~(\ref{eq:kast_formula}). Define 
    \begin{equation}
        \label{eq:istar_formula}
        i^\ast := \underset{i}{\argmax} \ i P_{ii}.
    \end{equation}
    We call the submatrix indexed by $[i^\ast] \times [i^\ast]$ the \emph{core submatrix}. Under the monotone Assumption~\ref{assum:mono}, for all entries $(i,j)$ in the core submatrix, the solution $k^\ast(i,j)$ coincides with $i^\ast$:
    \begin{lemma}
        \label{lem:core_submatrix_property}
        Under Assumption~\ref{assum:mono}, if $i \le i^\ast$ and $j \le i^\ast$, then $k^\ast (i,j) = i^\ast$.
    \end{lemma}
    \begin{IEEEproof}
        Note that $i^\ast P_{i^\ast i^\ast} \ge k P_{kk}$ for all $k$ by definition. 
        Additionally, we have $i^\ast P_{i^\ast i^\ast} \ge k P_{kk} \ge k P_{ik}$ for all $k < i$. For all $k < j$, we also have $i^\ast P_{i^\ast i^\ast} \ge k P_{kk} \ge k P_{kj}$.
    \end{IEEEproof}

\begin{figure}[htbp]
        \noindent \begin{centering}
        \begin{tabular}{ccc}
        \includegraphics[width=0.2\textwidth]{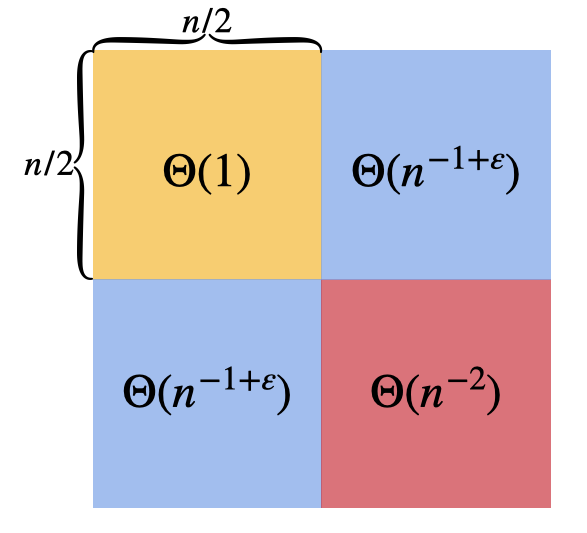} & ~~ & \includegraphics[width=0.17
        \textwidth]{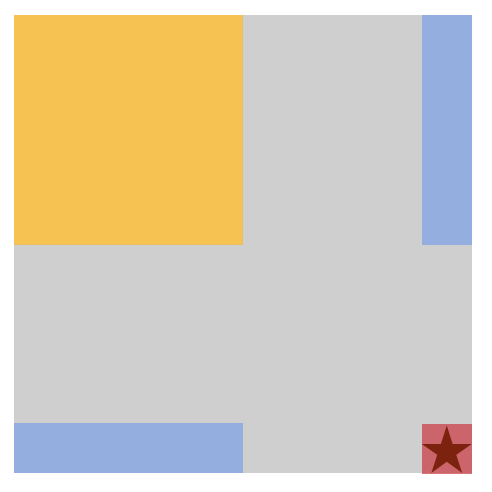}\tabularnewline
        \footnotesize{(a) Sturcture of $P$.} &  & \footnotesize{(b) Submatrix completion.} \tabularnewline
        \end{tabular}
        \par\end{centering}
        \caption{\label{fig:example} An example of using submatrix completion with monotone $P$.}
    \end{figure}

    For our example, suppose the probability matrix $P$ can be divided into four equal-size blocks, where the probabilities inside each block are the same up to constants, having the following structure:
    \begin{equation*}
        P_{ij} = 
        \begin{cases}
            \Theta(1) & i \le n/2, j \le n/2 \\
            \Theta(n^{-2}) & i > n/2, j > n/2 \\
            \Theta(n^{-1+\varepsilon}) & \text{otherwise},
        \end{cases}
    \end{equation*}
    for some $0<\varepsilon<1$. See Fig.~\ref{fig:example}(a) for an illustration of $P$.
    
    Since the smallest probability (red block) scales as $p_{\min} \asymp  n^{-2}$, existing entrywise error bound~\cite{Abbe2020entrywise} gives $\|\hat M - M^* \|_\infty = O(1/\sqrt{p_{\min}n})$, which is vacuous as $p_{\min}$ is so small.\footnote{We ignore logarithmic factors and dependence on the rank $r$.} In contrast, our submatrix completion algorithm achieves much better guarantees.
    In particular, it can be verified that the yellow block is the core submatrix. By Lemma~\ref{lem:core_submatrix_property}, to estimate entries inside the yellow block, our algorithm will apply SVT on this block itself and achieve an entrywise error bound of $O(1/\sqrt{n})$. 
    For each entry $(i,j)$ in the red block, our algorithm will use the submatrix $Y_{[n/2] \cup \{ i \}, [n/2] \cup \{ j \}}$ (see Fig.~\ref{fig:example}(b)) and achieve an $O(1/\sqrt{n^\varepsilon})$ error. Note that our error bounds are independent of $p_{\min}.$

    This example provides intuition for why it can be beneficial to  only use a subset of the observations for estimation. SVT applied to the entire matrix would try too hard to fit the (noisy) observations in the blue and red blocks with low sampling probabilities (and hence high variances). As a result, the estimation error for the high probability yellow block would be worse than using observations only  from this block.

\section{Theoretical Guarantees}
\label{sec:theoretical_guarantees}

    In this section, we present entry-specific error upper bounds for our submatrix completion algorithm coupled with SVT. We also derive entry-specific minimax lower bounds, which match the upper bound for structured $P$.

    \subsection{Upper bound for our algorithm}
    \label{subsec:upper_bound}
    
    We derive an error bound for estimating a specific entry $(i,j)$. \xxdelete{; the proof is deferred to Appendix~\ref{appen:proof_upper_bound}. }
    Set $p^\ast(i,j) = \min \{ P_{\max\{i,k^\ast\}, k^\ast}, P_{k^\ast, \max\{j,k^\ast\}} \}$ with $k^\ast \equiv k^*(i,j)$ being the solution to~(\ref{eq:kast_formula}). 
    Let $\hat M_{ij}$ be the estimate of $M^\ast_{ij}$ given by our submatrix completion algorithm. 
        \begin{theorem}
            \label{thm:upper_bound}
            Under Assumption~\ref{assum:mono}, with probability at least $1-\delta$, for each $(i,j)$ satisfying  $p^\ast(i,j) \ge \frac{c \log(n/\delta)}{k^\ast(i,j)}$, we have
            \begin{equation}
                \label{eq:upper_bound}
                \card{\hat M_{ij} - M^\ast_{ij}} \le C r \left( r + \sigma \right) \sqrt{\frac{\log^5(n/\delta)}{k^\ast (i,j) p^\ast(i,j)}}.
            \end{equation}
        \end{theorem}
    In the above upper bound, the denominator inside the square root is a function of the index $(i,j)$, where $k^*(i,j)$ is chosen precisely to maximize this denominator and hence optimize the error bound. As the denominator is increasing in the size of the submatrix but decreasing as the minimum probability decreases, our bound highlights the tension in the choice of the submatrix. A large submatrix may have a large size but a small minimum probability; a small submatrix has a small size but could have a larger minimum probability. This flexibility of choosing an appropriate submatrix enables us to obtain fine-grained error bounds that are specific to each entry. Compared with the uniform worst-case entrywise bound stated in Theorem~\ref{thm:uniformization_bound}, our bound is valid even when $\pmin$ does not meet the condition therein.
    Furthermore, our bound is able to capture the potential order-wise difference between the estimation quality of different entries, as demonstrated in the example from Section~\ref{sec:algorithm}.

    \subsection{Minimax lower bound}
    \label{subsec:lower_bound}

    We present an entry-specific minimax lower bound on the estimation error.  The following theorem\xxdelete{, proved in Appendix~\ref{appen:proof_lower_bound},} is valid for any sampling probability matrix~$P$.
    \begin{theorem}
        \label{thm:lower_bound}
        Fix $i\in [n]$ and $j \in [m]$. There exists an absolute constant $C>0$ such that
        \begin{equation}
                \label{eq:lower_bound_ij}
                \inf_{\hat M_{ij}} \sup_{M^\ast} \mathbb{E} \left[ \card{ \hat M_{ij} - M^\ast_{ij} } \right] \ge C \sigma \sqrt{\frac{r}{ \min \{\sum_{i'} P_{i'j}, \sum_{j'} P_{ij'} \} }} ,
        \end{equation}
        with probability at least $\frac12$.
        Here, the infimum is over all estimators of $M^*_{i,j}$, the supremum is over all rank-$r$ $M^\ast$, the expectation is w.r.t.\ the additive noise $E$, and the probability is w.r.t.\ the sampling mask $\Omega$.
    \end{theorem}

    We prove Theorem~\ref{thm:lower_bound} by reduction from noisy linear regression. Consider estimating the entry $M^\ast_{ij} = \langle a^*_i, b^*_j \rangle $. If the row latent factors $\{a_i^\ast \}_{i\in[n]}$ were known, then estimating $M^\ast_{ij}$ is the same as the linear regression problem of estimating the $j$-th column latent factor $b_j^\ast$ given noisy observations $Y_{i'j} = \langle a^*_{i'}, b^*_j \rangle + E_{i'j}$ for $i'\in[m]: \Omega_{i'j} = 1$; note that we have $\sum_{i'} P_{i'j}$ such observations in expectation. As the original problem is at least as hard as the regression problem, we can use standard minimax lower bounds for linear regression to derive a lower bound for our problem. A similar lower bound can be derived by assuming the column factors were known. Taking the larger of these two bounds proves Theorem~\ref{thm:lower_bound}.
    
    Perhaps surprisingly, for quite general settings of monotone $P$, the simple lower bound above matches the upper bound in Theorem~\ref{thm:upper_bound} (up to logarithmic factors), in which case our algorithm is information-theoretically optimal for estimating each entry. We discuss such a setting below.

    \subsection{Example: block-structured $P$}
    \label{subsec:block_P}
        
        We discuss a generalization of the example from Section~\ref{sec:example}. 
        Suppose $P$ can be partitioned into $4$ blocks: \footnote{This example can be generalized to $P$ with $O(1)$ blocks.} $ P = 
            \begin{pmatrix}
                Q_{11} & Q_{12} \\
                Q_{21} & Q_{22}
            \end{pmatrix},$
        where $Q_{11} \in \R^{n_1\times n_1}, Q_{12} \in \R^{n_1 \times n_2}, Q_{21} \in \R^{n_2 \times n_1}$, $Q_{22} \in \R^{n_2 \times n_2}$, and $n_1 + n_2 = n$. 
        Inside each block, the probabilities are of the same order but can be otherwise different. Let the minimum probabilities of the $4$ blocks be $q_{11}, q_{12}, q_{21}, q_{22}$. Assume the probabilities satisfy $ n_1 q_{11} \gtrsim n q_{12}$, $ n_1 q_{11} \gtrsim n q_{21}$, $n_1 q_{12} \gtrsim n q_{22}$, and $n_1 q_{21} \gtrsim n q_{22}$. (This assumption is satisfied when, for example, $P$ is monotone and $n_1 \gtrsim n_2$.)
        One may verify that for estimating entry $(i,j)$, our algorithm will pick the submatrix $Y_{[n_1]\cup\{i\}, [n_1]\cup\{j\}}$. Applying Theorems~\ref{thm:upper_bound} and~\ref{thm:lower_bound} to each block, we obtain the following matching upper and lower bounds (omitting log factors):\footnote{We impose the mild assumption $\min\{q_{11}, q_{12}, q_{21} \} \gtrsim \frac{1}{n_1}$, so that the problem is non-trivial with at least one observation in each row/column.} 
        
        \begin{enumerate}
            \item When $ i \le n_1 , j \le n_1$, the upper bound is $1 / \sqrt{n_1 q_{11}}$, and the lower bound is $ 1 / \sqrt{n_1 q_{11} + n_2 q_{12}}$.
            \item When $i \le n_1 , j > n_1$, the upper bound is $1 / \sqrt{n_1 q_{12}}$, and the lower bound is $ 1 / \sqrt{n_1 q_{12} + n_2 q_{22}}$.
            \item When $i > n_1 , j \le n_1$, the upper bound is $1 / \sqrt{n_1 q_{21}}$, and the lower bound is $ 1 / \sqrt{n_1 q_{21} + n_2 q_{22}}$.
            \item When $i > n_1, j > n_2$, the upper bound is $ 1 / \sqrt{n_1 \min\{q_{12}, q_{21}\}}$, and the lower bound is $ 1 / \sqrt{n_1 \min\{q_{12},q_{21}\} + n_2 q_{22}}$.
        \end{enumerate}

        

    \section{Key ideas of proof}
    \label{sec:uniform_to_non-uniform}

    In this section, we present the tools for proving the upper bound in Theorem~\ref{thm:upper_bound}. The high-level idea is to apply the entrywise error bound for SVT from~\cite{Abbe2020entrywise} to the submatrix chosen by our algorithm. To do so, their result needs to be adapted to the setting where the submatrix may have non-uniform sampling probabilities and one entry (the target entry to be estimated) may have an arbitrarily small probability. 

    We consider estimating a deterministic rank-$r$ matrix $A^{\ast}\in\R^{n \times m}$ given noisy observations $Y = \Omega \circ (A^\ast + E)$, where the mask $\Omega$ and noise $E$ are the same as before. \xxedit{Here, $A^\ast$ can be either the whole matrix $M^\ast$ introduced earlier or a submatrix of $M^\ast$. }
    Let the rank-$r$ SVD of $A^\ast$ be $A^{\ast} = U^{\ast}\Sigma^{\ast}V^{\ast \top}$. Define  $\kappa=\frac{\sigma_{1}^{\ast}}{\sigma_{r}^{\ast}}$ and $\eta=\left(\norm{U^{\ast}}_{2\to\infty}\vee\norm{V^{\ast}}_{2\to\infty}\right)$. Recall that the SVT algorithm forms the rescaled matrix $A_{ij} = (Y_{ij} / P_{ij})$ and computes the rank-$r$ truncated SVD $U\Sigma V^\top$ of $A$.

    \subsection{Guarantee for SVT with non-uniform observations}
    \label{subsec:uniformization}
    
    Leveraging the results from~\cite{Abbe2020entrywise}, we establish the following entrywise error bound for SVT under non-uniform sampling probabilities $P$. Let $\pmin := \min_{i,j} P_{ij}$.
        
        \begin{theorem}
             \label{thm:uniformization_bound}
            Suppose 
            $\pmin \ge \frac{c \log (n/\delta)}{n+m}$ 
            and 
            $\kappa\frac{\left(\left\Vert A^{\ast}\right\Vert _{\infty}+\sigma\right)}{\sigma_{r}^{\ast}}\sqrt{\frac{ (n+m)\log (n/\delta)}{\pmin}} \le 1$ for some $\delta>0$.
            With probability at least $1-\delta$, we have
            \begin{align}
                \norm{{U}{\Sigma}{V}^{\top} \!-\! A^{\ast}}_{\infty}  
                \le C\eta^{2}\kappa^{4}\left(\left\Vert A^{\ast}\right\Vert _{\infty} \!+\! \sigma\right)\sqrt{\frac{ (n+m) \log (n/\delta)}{\pmin}}.\label{eq:uniformization_bound}
            \end{align}
        \end{theorem}

    The bound~(\ref{eq:uniformization_bound}) depends on the smallest sampling probability $\pmin$. This bound becomes vacuous when just a single entry $(i,j)$ has a very small sampling probability $P_{ij}$. We next present an improved bound, which is unaffected by a few entries with small probabilities. This improvement plays a crucial role in proving our main Theorem~\ref{thm:upper_bound}.

    \subsection{Improved bound under a few small probabilities}
    \label{subsec:choose_entries}

        Let $s$ be a non-negative integer.  Let $p_{(1)} \ge p_{(2)} \ge \cdots \ge p_{(nm)}$ denote the probabilities $\{P_{ij}\}$ sorted in descending order. Note that $p_{(nm-s)}$ is the $(s+1)$-th smallest value in $\{P_{ij}\}$ and in particular $p_{(nm)} = \pmin$. The following theorem gives an error bound that only depends on $p_{(nm-s)}$. \xxedit{In order for the argument to hold, we need to slightly modify the way we rescale the observation matrix $Y$. Suppose $(i_{nm-s'}, j_{nm-s'})$ indicates the position of the probability $p_{(nm-s')}$ for $0 \le s' \le s-1$. Let $A_{i_{nm-s'}, j_{nm-s'}} = 2 Y_{i_{nm-s'}, j_{nm-s'}}$, where we replace the probability $p_{nm-s'}$ in the denominator with $\frac12$, enabling us to ignore probabilities smaller than $p_{(nm-s)}$.}
        
        \begin{theorem}
            \label{thm:choose_entries_bound}
            Suppose $p_{(nm-s)} \ge \frac{c \log (n/\delta) }{n+m}$ and $\kappa\frac{\left(\left\Vert A^{\ast}\right\Vert _{\infty}+\sigma\right)}{\sigma_{r}^{\ast}}\sqrt{\frac{ (n+m)\log (n/\delta)}{ p_{(nm-s)}}} \le 1$ for some $\delta>0$.
            With probability at least $1-\delta$, we have
            \begin{align}
                \norm{{U}{\Sigma}{V}^{\top} \!\!-\! A^{\ast}}_{\infty}  
                \le C \eta^{2}\kappa^{4}\left(\left\Vert A^{\ast}\right\Vert _{\infty} \!\!+\! \sigma\right)
                \sqrt{\frac{(n \!+\! m) (s \!+\! \log(n/\delta) )}{p_{(nm-s)}}}.
            \label{eq:choose_entries_bound}
            \end{align}
        \end{theorem}

        Compared with the bound~\eqref{eq:uniformization_bound}, the denominator on the right hand side of~\eqref{eq:choose_entries_bound} improves from $p_{(nm)}$ to $p_{(nm-s)}$, at the cost of the numerator increasing by $s$. This cost is negligible whenever $s = O(\log(n/\delta))$. To see the benefit, consider applying Theorem~\ref{thm:choose_entries_bound} with $s=1$ to the submatrix in Fig.~\ref{fig:example}(b), for which we obtain an error bound that depends on $p_{(nm-1)} = \Theta(n^{-1+\epsilon})$ instead of $p_{\min} = \Theta(n^{-2})$.

\section{Numerical experiments}
\label{sec:numerical_experiments}

    In this section, we numerically evaluate our algorithm. We compare two algorithms: (i) our algorithm SVT-sub, which applies SVT to submatrices, and (ii) SVT-whole, which applies SVT to the entire matrix.
    We consider two monotone probability matrices $P$. For each case, we randomly generate a $100$-by-$100$ signal matrix $M^\ast$ of rank $r=2$ according to the latent variable model in Section~\ref{subsec:latent_variable_model} with noise standard deviation $\sigma=0.1$. This is repeated for $100$ trials. We record the error $e_{ij}^{\text{sub}}$ and $e_{ij}^{\text{whole}}$ for estimating each entry $(i,j)$ using SVT-sub and SVT-whole, respectively, averaged over 100 trials.

    \subsection{Block-constant $P$ matrix}
    \label{subsec:block-constant}
\begin{figure}[htbp]
            \noindent \begin{centering}
            \begin{tabular}{cc}
            \includegraphics[width=0.18\textwidth]{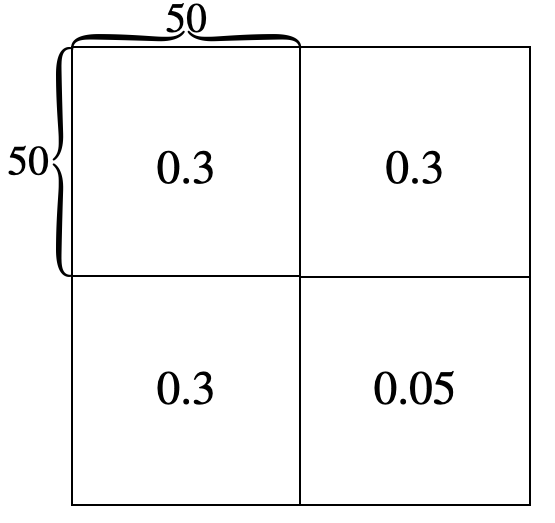} & \includegraphics[width=0.21\textwidth]{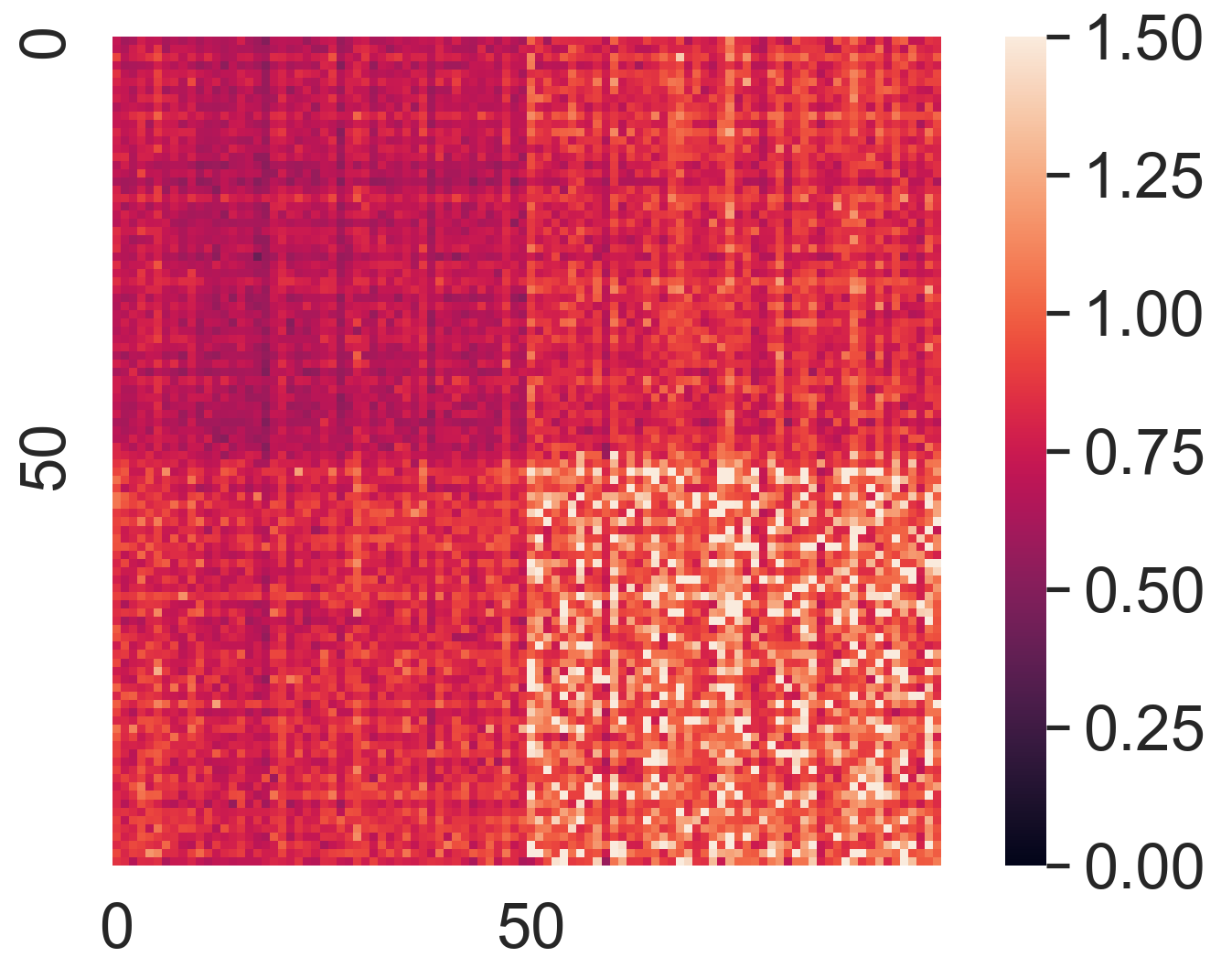}
            \tabularnewline
            (a) $2$-by-$2$ block-constant $P$. &  (b) Error by SVT-whole.  
            \tabularnewline
            \includegraphics[width=0.21\textwidth]{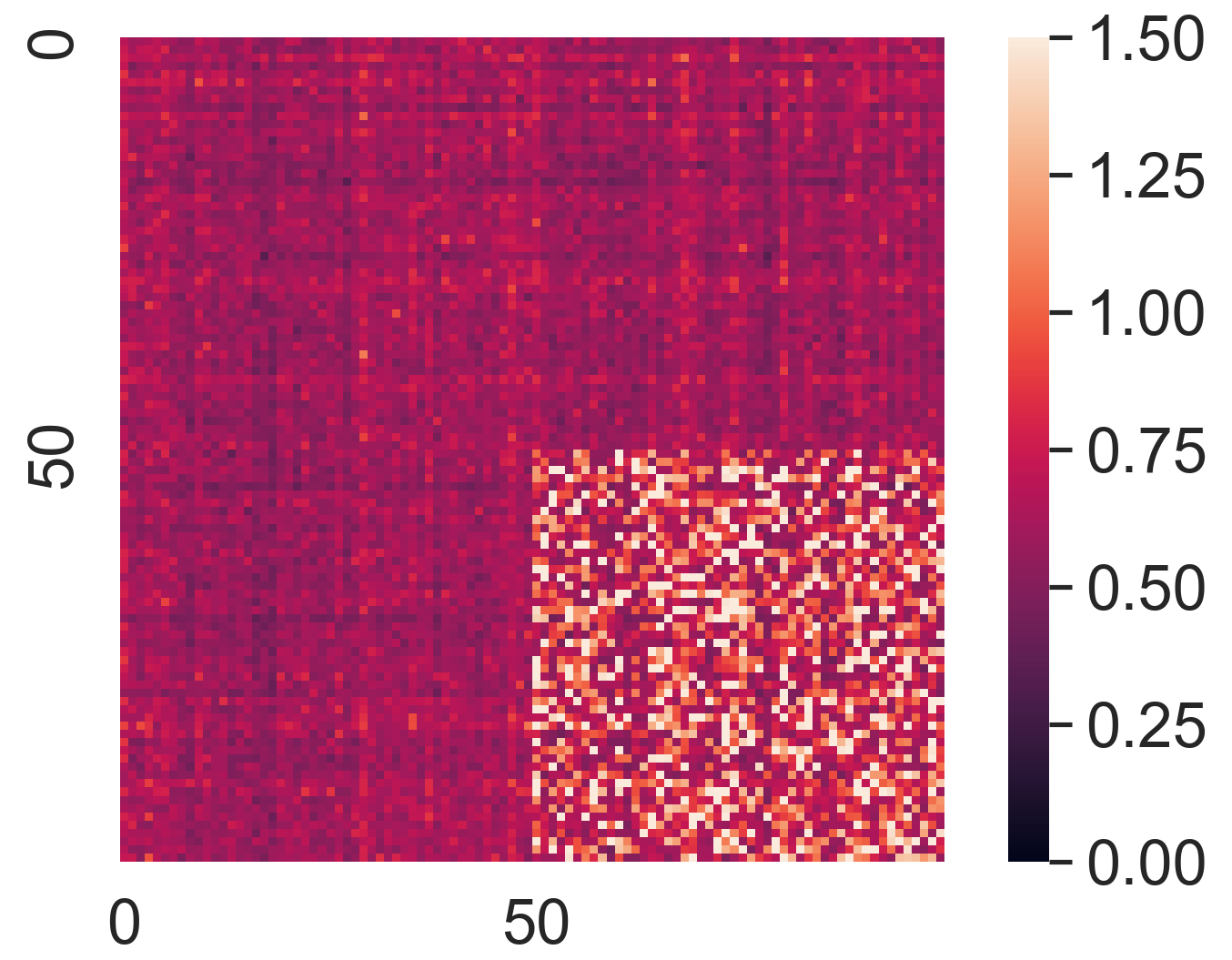} & \includegraphics[width=0.21\textwidth]{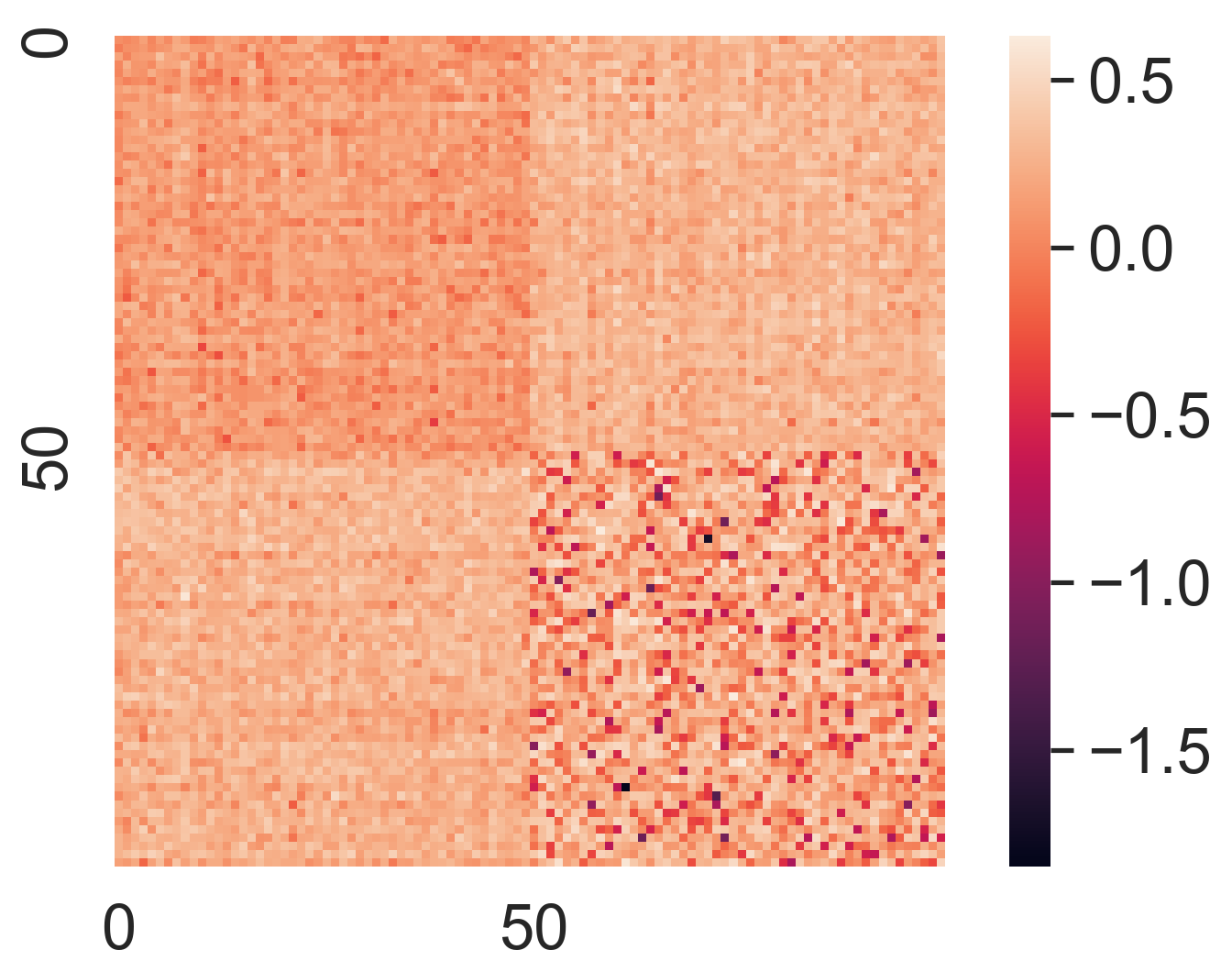}
            \tabularnewline
            (c) Error by SVT-sub. &  (d) Relative improvement.
            \end{tabular}
            \par\end{centering}
            \caption{\label{fig:block-constant} Heatmaps for Subsection~\ref{subsec:block-constant}.}
        \end{figure}

        We first consider a block-constant $P \in [0,1]^{100 \times 100}$ with 
        \[
            P_{ij} = 
            \begin{cases}
                0.3 & i \le 50 \text{ or } j \le 50, \\
                0.05 & i > 50 \text{ and } j > 50,
            \end{cases}
        \]
        which is visualized in Fig.~\ref{fig:block-constant}(a). 
        In this case, our algorithm SVT-sub uses the submatrix $M_{[50] \cup \{i\}, [50] \cup \{j\}}$ to estimate each entry $(i,j)$. We plot the heatmaps of the errors $e_{ij}^{\text{sub}}$ and $e_{ij}^{\text{whole}}$ in Fig.~\ref{fig:block-constant}(b) and~(c), respectively. We observe that SVT-sub achieves a smaller error, especially in the three $0.3$ blocks. We further compute the \emph{relative improvement} for estimating each entry $(i,j)$, defined as $(e_{ij}^{\text{whole}} - e_{ij}^{\text{sub}})/e_{ij}^{\text{whole}},$
        which represents the percentage of improvement of SVT-sub over SVT-whole. 
        We plot the relative improvements in Fig.~\ref{fig:block-constant}(d), which shows that the most substantial improvement happens in the two off-diagonal blocks. 
        In particular, SVT-sub improves upon SVT-whole by $12.7\%$ on average over the top-left block, by $21.3\%$ over the two off-diagonal blocks, and by $14.5\%$ over the bottom-right block.  
        We also plot the histogram of the relative improvements in Fig.~\ref{fig:histgrams}(a), showing a strong trend of positive improvement.

    \subsection{Rank-one $P$ matrix}
    \label{subsec:rank-1}
    
\begin{figure}[htbp]
            \noindent \begin{centering}
            \begin{tabular}{cc}
            \includegraphics[width=0.21\textwidth]{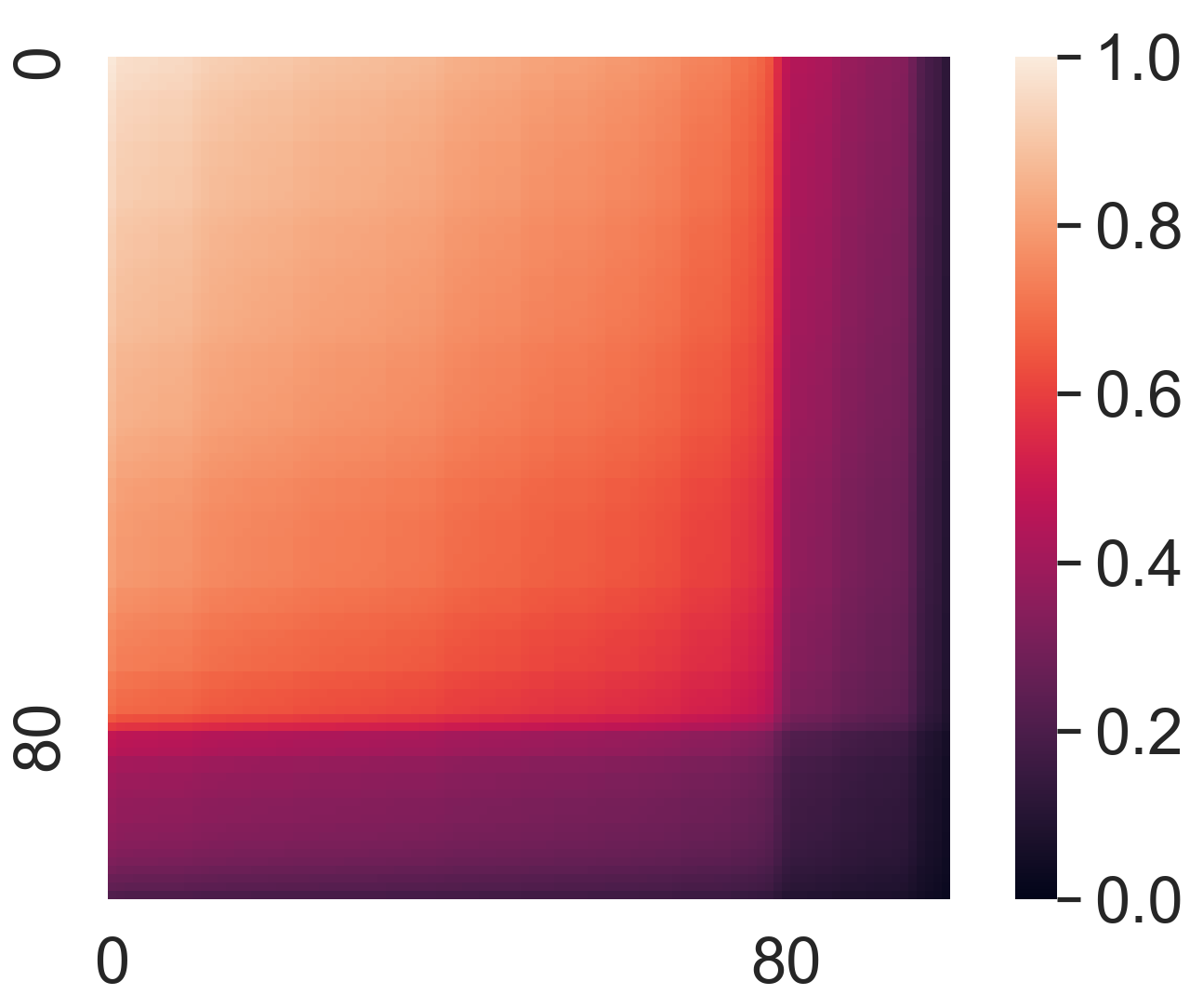} & \includegraphics[width=0.21\textwidth]{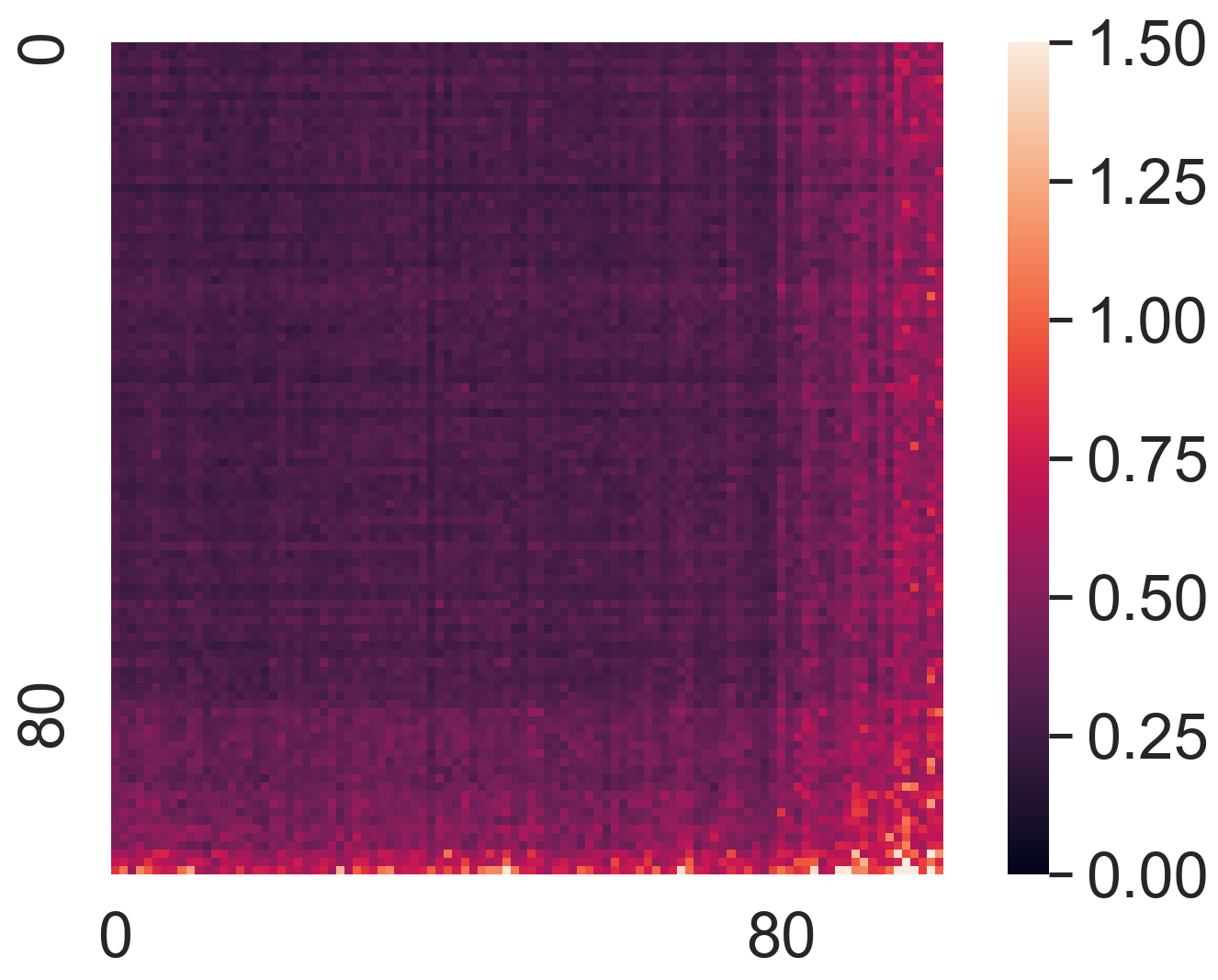}
            \tabularnewline
            (a) Rank-$1$ $P$. &  (b) Error by SVT-whole.  
            \tabularnewline
            \includegraphics[width=0.21\textwidth]{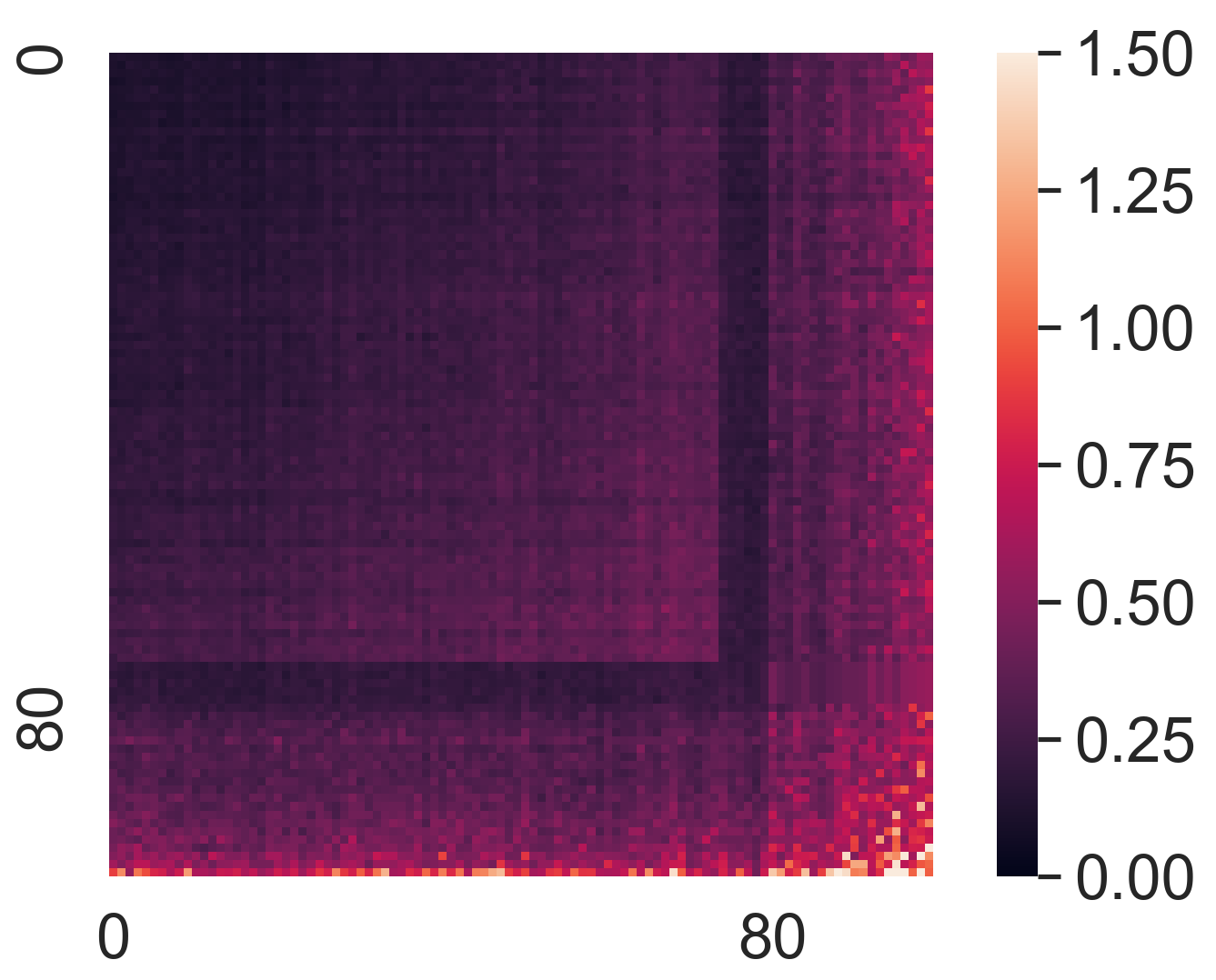} & \includegraphics[width=0.21\textwidth]{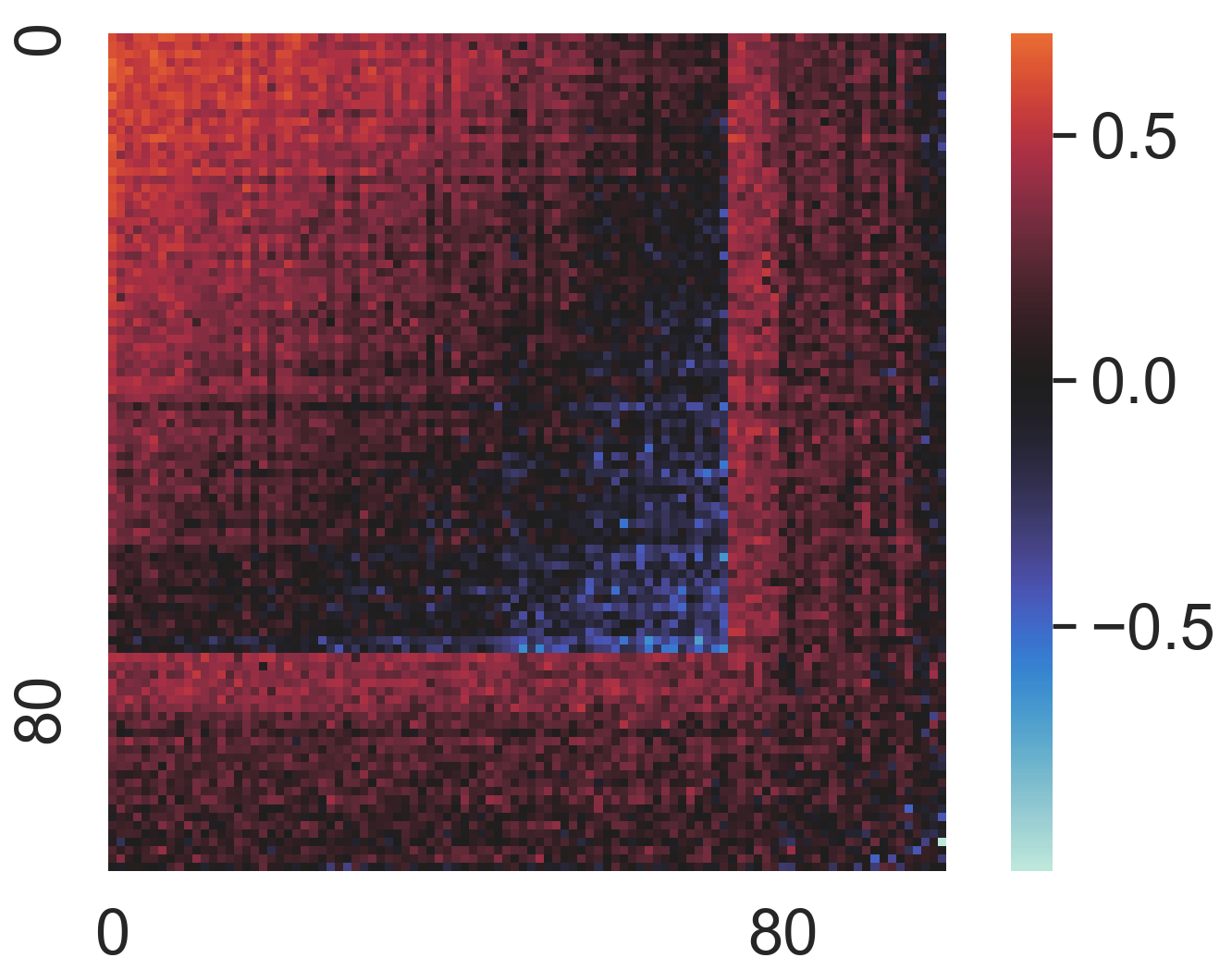}
            \tabularnewline
            (c) Error by SVT-sub. &  (d) Relative improvement.
            \end{tabular}
            \par\end{centering}
            \caption{\label{fig:rank-1} Heatmaps for Subsection~\ref{subsec:rank-1}.}
        \end{figure}

\begin{figure}[htbp]
            \noindent \begin{centering}
            \begin{tabular}{cc}
            
            \includegraphics[width=0.23\textwidth]{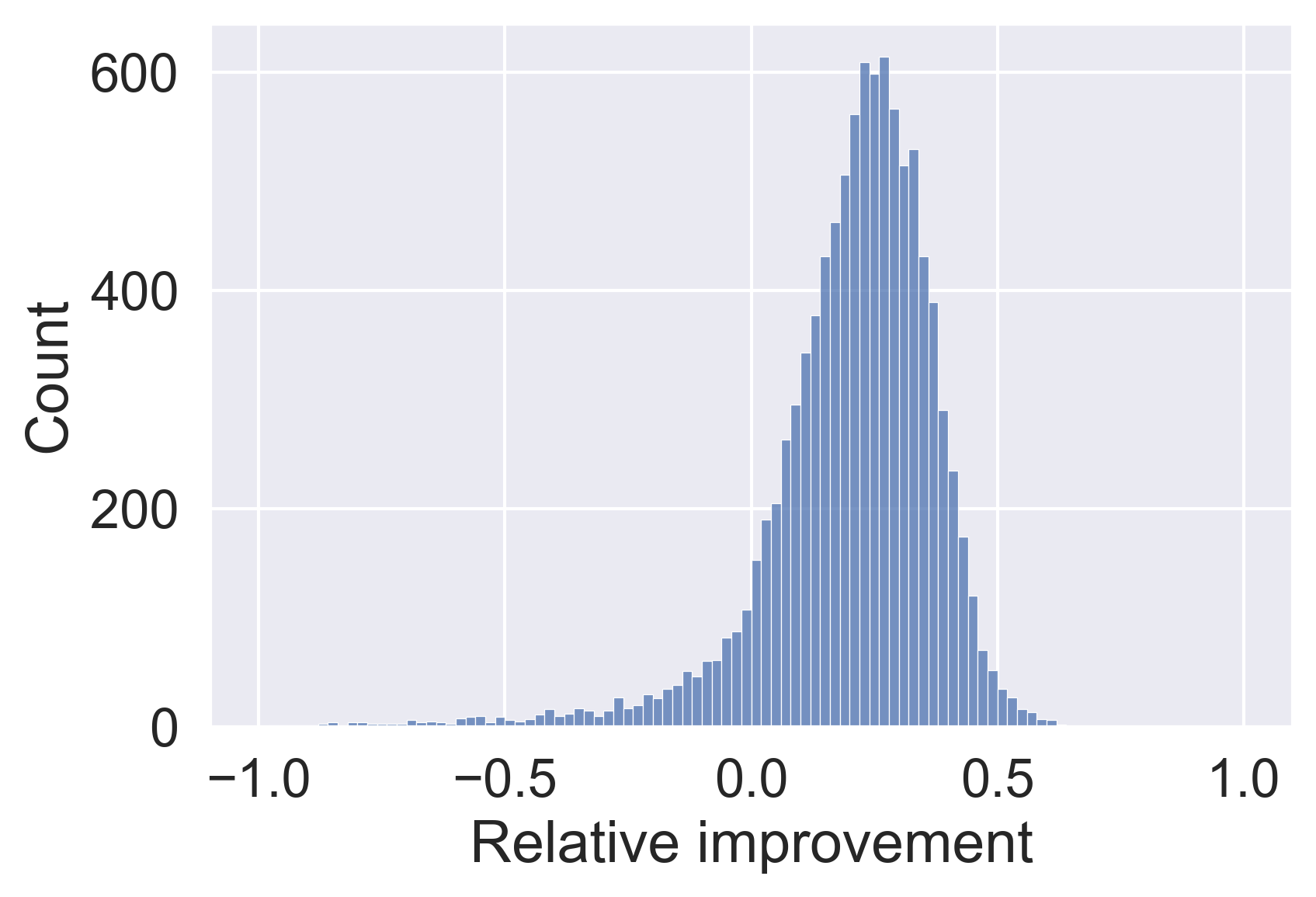} & 
            \includegraphics[width=0.23\textwidth]{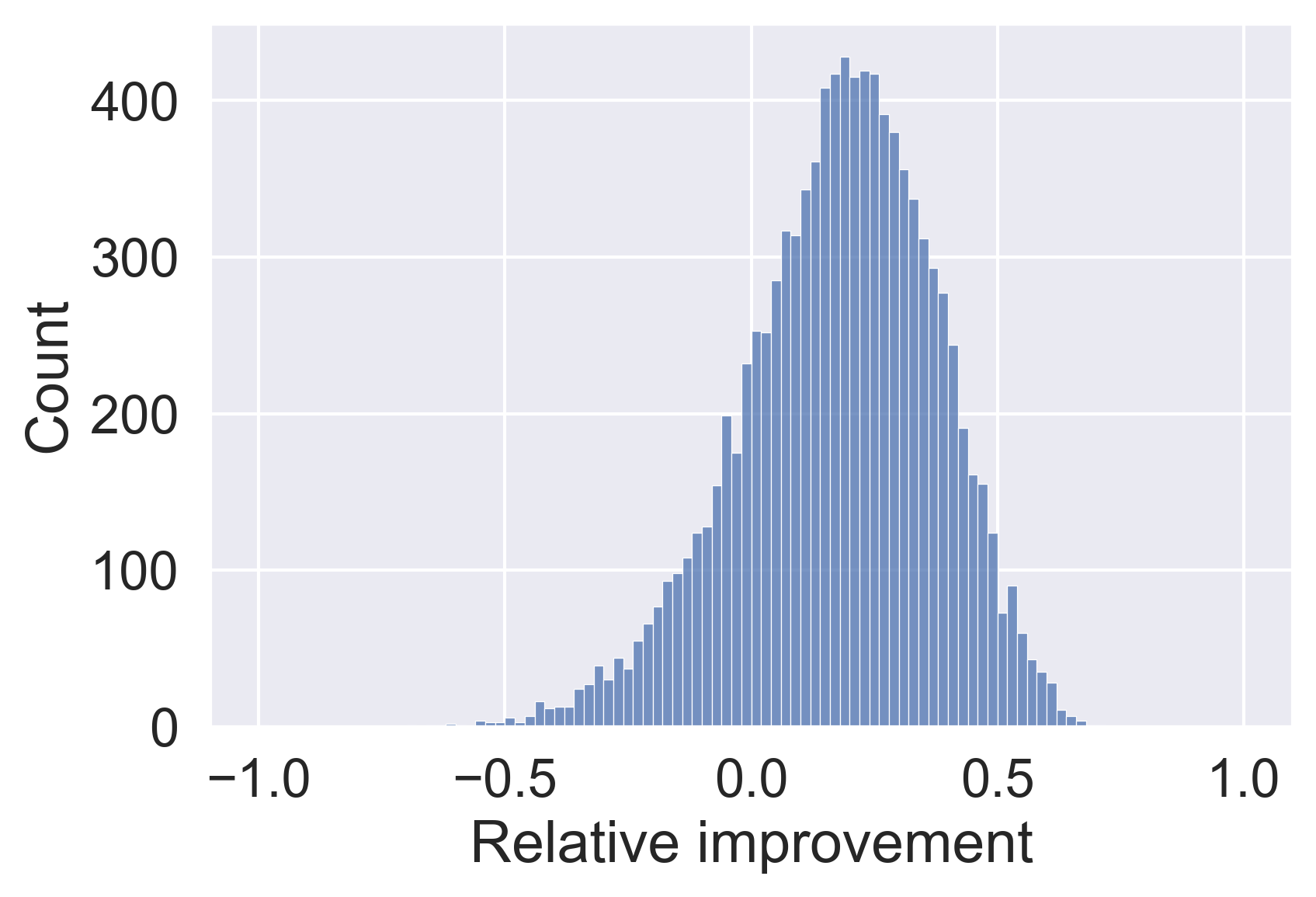} 
             \\
             \footnotesize{(a) For Subsection~\ref{subsec:block-constant}.} & \footnotesize{(b)  For Subsection~\ref{subsec:rank-1}. }
            \end{tabular}
            \par\end{centering}
            \caption{\label{fig:histgrams} Histograms of relative improvement.}
        \end{figure}

        We consider a rank-one $P = ab^\top$, where $a$ and $b$ are sampled randomly from the same distribution. In particular, for $i \le 80$, we have $a_i \sim  0.5 \cdot \text{Beta}(5,2) + 0.5$; for $i > 80$, we have $a_i \sim  0.5\cdot \text{Beta}(5,2)$. We sort the entries of $a$ and $b$ in descending order for better visualization.
        One realization of $P$ is shown in Fig~\ref{fig:rank-1}(a), in which we observe a clear drop in the probabilities near the $80$th row/column. In this realization, the largest and smallest values of $P_{ij}$ are $0.989$ and $0.025$, respectively, hence the sampling probabilities are highly non-uniform. The core matrix (see Section~\ref{sec:example}) obtained by solving~(\ref{eq:istar_formula}) is a $74$-by-$74$ submatrix.
        We plot the heatmaps of the entrywise error  $e_{ij}^{\text{sub}}$ and $e_{ij}^{\text{whole}}$ in Fig.~\ref{fig:rank-1}(b) and~(c), respectively, as well as the relative improvements in Fig.~\ref{fig:rank-1}(d) and Fig.~\ref{fig:histgrams}(b). We observe that the majority of the entries benefit from using our SVT-sub algorithm. On average, the relative improvement is $17.7\%$.

\section{Discussion}
    We propose a submatrix completion algorithm, which handpicks a submatrix for estimating a specific entry based on the sampling probabilities and then applies the matrix estimation subroutine to the selected submatrix. Using SVT as the subroutine, we establish entry-specific upper bound and minimax lower bound on the estimation error. Under certain sampling probability patterns, the upper and lower bounds match up to log factors. We also present numerical experiments that demonstrate the benefit of our algorithm. 
    Future directions include combining our algorithm with other matrix estimation algorithms, as well as extending the results to more general probability patterns, such as those that are not monotone globally but may satisfy local monotonicity.

\textbf{Acknowledgement:}  C.\ Yu is partially supported by NSF grants CCF-1948256 and CNS-1955997, AFOSR grant FA9550-23-1-0301, and by an Intel Rising Stars award. Y.\ Chen is partially supported by NSF grants CCF-1704828 and CCF-2047910.

\clearpage
\bibliographystyle{plain}
\bibliography{ref}

\clearpage
\appendices

\section{Technical Lemmas}
\label{appen:technical_lemmas}

        Before we present the technical lemmas, we define some notations.
        For a vector $x \in \R^n$, we define $\twonorm{x} = \sqrt{\sum_{i \in [n]} x_i^2}$ and $\inftynorm{x} = \max_{i \in [n]} \card{x_i}$. 
        For a matrix $M \in \R^{n \times m}$, let $M_{i\cdot}$ denote its $i$-th row and $M_{\cdot j}$ its $j$-th column. 
        Let the operator norm of matrix $M \in \R^{n \times m}$ be $\opnorm{M} = \max_{\twonorm{x}=1, \twonorm{y}=1} x^\top M y$ and the $2 \to \infty$ norm be $\twotoinftynorm{M} = \max_{\twonorm{x}=1} \inftynorm{Mx} = \max_i \twonorm{M_{i\cdot}}$. 

        Recall that in the latent variable model introduced in Subsection~\ref{subsec:latent_variable_model}, the signal matrix $M^\ast \in \R^{n \times m}$ is generated via $ M^\ast_{ij} = \langle a_i^\ast, b_j^\ast \rangle$.
        The latent variables are standard Gaussians: $a_i^\ast \widesim{\text{i.i.d.}} N(0,I_r)$ and  $b_j ^\ast \widesim{\text{i.i.d.}} N(0,I_r)$, independently. 
        We introduce the following lemmas to show the signal matrix is low-rank and bounded with high probability, the proof of which is deferred to Appendix~\ref{appen:proof_low-rank_signal} and Appendix~\ref{appen:proof_bounded_signal}. 

        \begin{lemma}[Low-rank signal]
            \label{lem:low-rank_signal}
            Assume that $\delta > 0$ satisfy $r + \log (4 / \delta) \le \frac{1}{C} \min \{ n,m\}$, for a sufficiently large absolute constant $C>0$.
            Then, with probability at least $1-\delta$, matrix $M^\ast$ is rank-$r$.
        \end{lemma}

        \begin{lemma}[Bounded signal]
            \label{lem:bounded_signal}
            With probability at least $1-\delta$, we have
            \begin{equation}
                \inftynorm{M^\ast} \le 2r \log \frac{nmr^2}{\delta}.
            \end{equation}
        \end{lemma}

        The next lemma shows that the signal matrix $M^\ast$ is incoherent and well-conditioned, with its proof in Appendix~\ref{appen:proof_incoherence_condition}.

        \begin{lemma}[Incoherence \& condition number guarantee]
            \label{lem:incoherence_condition}
            Let the SVD of $M^\ast$ be $M^\ast = U^\ast \Sigma^\ast V^{\ast \top}$. There exists a sufficiently large absolute constant $C>0$ such that with probability at least $1-\delta$, we have 
            \begin{equation}
                \label{eq:condition_M}
                \kappa(M^\ast) \le \frac{ 1 + C \sqrt \frac{r + \log(2(n+m+2)/\delta)}{\min\{n,m\}}}{ 1 - C \sqrt \frac{r + \log(4(n+m+2)/\delta)}{\min\{n,m\}}}.
            \end{equation}
            Furthermore, $M^\ast$ is incoherent:
            \begin{equation}
                \label{eq:incoherence_UV}
                \begin{split}
                    \norm{U^\ast}_{2 \to \infty} &\le C \frac {\sqrt{r} + \sqrt{\log(2(n+m+2) / \delta)}}{\sqrt{n}} \\
                \norm{V^\ast}_{2 \to \infty} &\le C \frac {\sqrt{r} + \sqrt{\log(2(n+m+2) / \delta)}}{\sqrt{m}}.
                \end{split}
            \end{equation}
        \end{lemma}

        

    \subsection{Proof of Lemma~\ref{lem:low-rank_signal}}
        \label{appen:proof_low-rank_signal}
        Write $A = \begin{pmatrix}
            a_1^\top \\
            a_2^\top \\
            \vdots \\
            a_n^\top
        \end{pmatrix} \in \R^{n \times r}$ 
        and 
        $B = \begin{pmatrix}
            b_1^\top \\
            b_2^\top \\
            \vdots \\
            b_m^\top
        \end{pmatrix}
        \in \R^{m \times r}.$
        We rescale the Gaussian vectors and define
        $U = \frac{1}{\sqrt{n}} A$ and $V = \frac{1}{\sqrt{m}} B$. Then, we have
        $U^\top U = \frac1n \sum_{i \in [n]} a_i a_i^\top$ and $V^\top V = \frac1m \sum_{i \in [m]} b_i b_i^\top$.
        By a standard argument of covariance estimation~\cite{wainwright2019high-dim_stats}, there exists a sufficiently large absolute constant $C>0$ such that
        \begin{equation}
            \label{eq:U_concentration}
            \opnorm{U^\top U - I} \le C \sqrt \frac{r + \log(4 / \delta)}{n},
        \end{equation}
        with probability at least $1-\delta/2$. By the same argument, we have
        \begin{equation}
            \label{eq:V_concentration}
            \opnorm{V^\top V - I} \le C \sqrt \frac{r + \log(4 / \delta)}{m},
        \end{equation}
        with probability at least $1-\delta/2$. Applying the union bound on the aforementioned two events yields that with probability at least $1-\delta$, we have
        $\rank(U) = r$ and $\rank(V) = r$, since all singular values of $U$ and $V$ are concentrated around $1$, by Weyl's inequality.
        Because $M^* = \sqrt{nm} UV^\top$,  we immediately have $\rank(M) \le \min \{ \rank(U), \rank(V) \} = r$.
        Finally, invoking Sylvester's rank inequality, we get $\rank (M^*) \ge \rank(U) + \rank(V) - r = r$. As a result, $\rank(M^*)=r$ with probability at least $1-\delta$.

    \subsection{Proof of Lemma~\ref{lem:bounded_signal}}
        \label{appen:proof_bounded_signal}
        Note that for each $a_{ik}^\ast \sim N(0,1)$ and $b_{j\ell }^\ast \sim N(0,1)$, we have the Gaussian tail bound
        \begin{align}
            \label{eq:ai_gaussian_tail}
            \mathbb{P} ( \card{a_{ik}^\ast} \ge t) \le \exp \left( -\frac{t^2}{2} \right).
        \end{align}
        We then derive the tail bound on the maximum of $\card{a_{ik}^\ast}$ and $\card{b_{j\ell}^\ast}$ as
        \begin{align*}
            & \mathbb{P} ( \max \{ \max_{i\in[n], k \in [r]} \card{a_{ik}^\ast}, \max_{j \in [m], \ell  \in [r]} \card{b_{j\ell }^\ast} \} \ge t) \\ 
            \le &\mathbb{P} (  \{ \cup_{i,k} \{ \card{a_{ik}^\ast} \ge t \} \} \cup \{\cup_{j,\ell} \{ \card{b_{j\ell}^\ast} \ge t \}  \} ) \\
            \le & \sum_{i,k}  \mathbb{P} ( \card{a_{ik}^\ast} \ge t)  + \sum_{j, \ell} \mathbb{P} ( \card{b_{j \ell}^\ast} \ge t)\\
            \le & nmr^2 \exp \left( -\frac{t^2}{2} \right).
        \end{align*}
        Let $t=\sqrt{2 \log \frac{nmr^2}{\delta}}$ and we get 
        \[
             \mathbb{P} \left(  \max \{ \max_{i\in[n], k \in [r]} \card{a_{ik}^\ast}, \max_{j \in [m], \ell  \in [r]} \card{b_{j\ell }^\ast} \} \ge \sqrt{2 \log \frac{nmr^2}{\delta}} \right) \le \delta.
        \]
        Hence, we conclude that $\max_{ij} \card{M^\ast_{ij}} \le 2r \log \frac{nmr^2}{\delta}$ with probability at least $1-\delta$.

    \subsection{Proof of Lemma~\ref{lem:incoherence_condition}}
        \label{appen:proof_incoherence_condition}
        Continuing from the previous proof, we have $M^* = \sqrt{nm} UV^\top $ where $U$ and $V$ satisfy~(\ref{eq:U_concentration}) and~(\ref{eq:V_concentration}). 
            Let the (reduced) QR decomposition of $U$ and $V$ be
            \begin{align*}
                U & = Q^U R^U \\
                V & = Q^V R^V.
            \end{align*}
            Plugging the above decomposition into $M^* = \sqrt{nm} UV^\top $, we have
            \begin{align*}
                M^* &= \sqrt{nm} Q^U (R^U R^{V \top}) Q^{V \top} \\
                &= \sqrt{nm} Q^U (P^R \Sigma^R Q^{R\top}) Q^{V \top},
            \end{align*}
            where in the last line, we further decompose matrix $R^U R^{V \top}$ by its singular vectors and singular values. Rewriting the last line, we get
            \[
                M^* = \sqrt{nm} (Q^U P^R) \Sigma^R (Q^V Q^R)^\top = U^* \Sigma^* V^{* \top},
            \]
            where 
            Note that both $Q^U P^R$ and $Q^V Q^R$ are orthogonal.
            Hence, we derive the two-to-infinity norm bound of $U^*$ and $V^*$ by
            \begin{align*}
                \norm{U^*}_{2 \to \infty}  & \le \norm{Q^U}_{2 \to \infty} \opnorm{P^R} =  \norm{Q^U}_{2 \to \infty} \\
                \norm{V^*}_{2 \to \infty}  & \le \norm{Q^V}_{2 \to \infty} \opnorm{Q^R} =  \norm{Q^V}_{2 \to \infty},
            \end{align*}
            using the fact that orthogonal matrices are norm-preserving.
            Note that $Q^U$ and $Q^V$ are comprised of (right) singular vectors of $U$ and $V$, respectively. We can bound the two-to-infinity norm of $Q^U$ by the two-to-infinity norm of $U$ and the inverse of the largest singular value of $U$:
            \begin{align*}
                 \norm{Q^U}_{2 \to \infty} & =  \norm{UR^{U, -1}}_{2 \to \infty} \\
                 & \le \norm{U}_{2 \to \infty} \opnorm{R^{U, -1}} \\
                 & = \norm{U}_{2 \to \infty} \opnorm{U}^{-1}.
            \end{align*}
            Similarly, we get
            \[
                \norm{Q^V}_{2 \to \infty} \le  \norm{V}_{2 \to \infty} \opnorm{V}^{-1}.
            \]
            Next, we condition on the union of several high-probability events to upper bound the norms of $U$ and $V$. Recall that for standard normal vectors of dimension $r$, its norm will concentrate around $\sqrt r$.
            There exists a sufficiently large absolute constant $C>0$ such that with probability at least $1-\delta$, the following is true. 
            \begin{align*}
                \opnorm{U^\top U - I} & \le C \sqrt \frac{r + \log(2(n+m+2)  / \delta)}{n} \\
                \opnorm{V^\top V - I} & \le C \sqrt  \frac{r + \log(2(n+m+2) / \delta)}{m} \\
                \norm{U}_{2 \to \infty} & \le \frac {\sqrt{r} + \sqrt{\log(2(n+m+2) / \delta)}}{\sqrt{n}} \\
                \norm{V}_{2 \to \infty} & \le \frac{\sqrt{r} + \sqrt{\log(2(n+m+2) / \delta)}}{\sqrt{m}}.
            \end{align*}
            Under the above event, we conclude that
            \begin{align*}
                 \norm{U^*}_{2 \to \infty}  & \lesssim  \norm{U}_{2 \to \infty}  \lesssim \sqrt{\frac{r}{n} } \\ 
                  \norm{V^*}_{2 \to \infty}  & \lesssim  \norm{V}_{2 \to \infty}  \lesssim \sqrt{\frac{r}{m} }. \\
            \end{align*}
            Finally, the condition number of $M$ can be upper bounded because 
            $\sigma_{\max}(UV^\top) \le \sigma_{\max}(U) \sigma_{\max}(V) = 1 + O(\sqrt{\frac{r}{n}}) + O(\sqrt{\frac{r}{m}}) $ and $\sigma_{\min} (UV^\top) \ge \sigma_{\min}(U) \sigma_{\min}(V) = 1 + O(\sqrt{\frac{r}{n}})+ O(\sqrt{\frac{r}{m}})$.

\section{Proofs for Section~\ref{sec:uniform_to_non-uniform}}

    \subsection{Proof of Theorem~\ref{thm:uniformization_bound}}
    \label{appen:proof_uniformization_bound}
        To apply Theorem~2.1 in~\cite{Abbe2020entrywise}, we follow their proof for the symmetric matrix completion and extend it to the general case via the ``symmetric dilation'' trick.
        Let $A^\ast \in \R^{n \times n}$ be symmetric, $A = (A_{ij})$ with $A_{ij} = (A_{ij}^\ast + E_{ij}) \Omega_{ij} / P_{ij}$ and  $\bar A = (\bar A_{ij})$  with $\bar A_{ij} = A^\ast_{ij} \Omega_{ij} / P_{ij}$. We check spectral norm concentration in Lemma~\ref{lem:spectral_norm_concentration} and row norm concentration in Lemma~\ref{lem:row_norm_concentration}.
        Assume $\pmin \ge \frac{c \log(n /\delta)}{n}$.
        
        \begin{lemma}
            \label{lem:spectral_norm_concentration}
            There exists a constant $C>0$ such that with probability at least $1-\delta$, 
            \begin{align*}
                \opnorm{\bar A - A^\ast } \le & C \inftynorm{A^\ast} \sqrt{\frac{n\log(n/\delta)}{\pmin}} \\
                \opnorm{A - \bar A} \le & C \sigma \sqrt{\frac{n\log(n/\delta)}{\pmin}}.
            \end{align*}
        \end{lemma}
        \begin{IEEEproof}
            WLOG, assume $\inftynorm{A^\ast}=1$.
            To prove the first inequality, we invoke the matrix Bernstein inequality~\cite{tropp2015introduction} and obtain
            \[
                \mathbb{P} \left( \opnorm{\bar A - A^\ast} \ge t \right) \le 2n \exp \left( \frac{-\pmin t^2/2}{n + t/3} \right).
            \]
            Setting $t = \sqrt{ \frac{2n \log (2n/\delta)}{\pmin} }$ yields the desired result.

            To prove the second inequality, we define $Z_{ij} = E_{ij} \ind_{\{ \card{E_{ij}} \le c_1 \sqrt{\log(n/\delta)}\}}$ and $\tilde A_{ij} =  (A_{ij}^\ast + Z_{ij}) \Omega_{ij} / P_{ij}$.  By applying matrix Bernstein's inequality, we get
            
            \begin{align*}
                \mathbb P \left( \opnorm{\tilde A - \bar A} \ge t \right) \le 2n \exp \left( \frac{-\pmin t^2/2}{n \sigma^2 + t \log(n/\delta) / 3} \right).
            \end{align*}
            
            By setting $t= \sigma \sqrt{\frac{n \log(n/\delta)}{\pmin}}$, we obtain
            \[
                \opnorm{\tilde A - \bar A} \lesssim \sigma \sqrt{\frac{n \log ( n/\delta)}{\pmin}},
            \]
            with probability at least $1-\delta$. 
            Using the standard Gaussian tail bound, we can also show that
            \begin{align*}
                \mathbb P (\tilde A - A \neq 0) & \le \cup_{i,j} \mathbb P( \card{\varepsilon_{ij}} \ge c_1 \sqrt{\log n/\delta}) \\
                & \le n^2 \exp \left(-\frac{c_1 \log (n/\delta)}{2} \right) \le \delta,
            \end{align*}
            provided $c_1$ is sufficiently large, which completes the proof.
        \end{IEEEproof}

        In the next lemma, we adopt similar notations as~\cite{Abbe2020entrywise} and define $\bar \varphi(x)$ and $\tilde \varphi (x)$ exactly the same. Our proof is not too different from theirs. So, we first check some inequalities in the non-uniform setting and then refer readers to their proof.
        \begin{lemma}
            \label{lem:row_norm_concentration}
            For any fixed $W \in \R^{n \times r}$ and $k \in [n]$, we have
            \begin{align*}
                \twonorm{(\bar A - A^\ast)_{k \cdot} W} \le & C \twotoinftynorm{W} \bar \varphi \left( \frac{\fnorm{W}}{\sqrt n \twotoinftynorm{W}} \right) \\
                \twonorm{(A - \bar A)_{k \cdot} W} \le & C \twotoinftynorm{W} \tilde \varphi \left( \frac{\fnorm{W}}{\sqrt n \twotoinftynorm{W}} \right) ,
            \end{align*}
            with probability at least $1-\delta$.
        \end{lemma}
        \begin{IEEEproof}
            Let $X_i = (\bar A - A^\ast)_{k i} W_{i\cdot}$ for all $i\in [n]$. We derive that
            \[
                \twonorm{X_i} \le \inftynorm{A^\ast} \twotoinftynorm{W} / P_{ki} \le \inftynorm{A^\ast} \twotoinftynorm{W} / \pmin
            \]
            and 
            \begin{align*}
                \mathbb E \twonorm{X_i}^2 & = \twonorm{W_{i\cdot}}^2 \mathbb E \bar A_{ki}^2 \\
                & \le \inftynorm{A^\ast}^2 \twonorm{W_{i\cdot}}^2 / P_{ki} \\
                & \le \inftynorm{A^\ast}^2 \twonorm{W_{i\cdot}}^2 / \pmin.
            \end{align*}
            Applying matrix Bernstein's inequality as in~\cite{Abbe2020entrywise}, we obtain the desired bound.

            To prove the second inequality, we define $S_k = \{ i \in [n] \colon \Omega_{ki} = 0\}$ as the observed entries in the $k$-th row. Then, we have
            \begin{align*}
                (A - \bar A)_{k\cdot} W  & = \sigma \sum_{i \in S_k} \frac{E_{ki}}{\sigma} \cdot \frac{W_{i\cdot}}{P_{ki}} \\
                &= \frac{\sigma}{\pmin}  \sum_{i \in S_k} \frac{E_{ki}}{\sigma} \cdot \frac{\pmin W_{i\cdot}}{P_{ki}}.
            \end{align*}
            Given $S_k$, we note that $\{E_{ki} / \sigma\}_{i \in S_k}$ are i.i.d.~$N(0,1)$.
            Additionally, we have
            \[
                \sum_{i \in S_k} \twonorm{\frac{\pmin W_{i\cdot}}{P_{ki}}}^2 \le  \sum_{i \in S_k} \twonorm{W_{i\cdot}}^2. 
            \]
            Furthermore, using Chernoff's bound, we can also derive
            \[
                \mathbb{P} ( \card{S_k} \ge 2 \sum_{k \in[n]} P_{ki} ) \le \delta, 
            \]
            since $\pmin \ge \frac{c \log (n/\delta) }{n}$. 
            The rest of the proof follows~\cite{Abbe2020entrywise}, where they first show the concentration of a matrix Gaussian sequence given $S_k$ and then invoke Chernoff's inequality to  control the cardinality of $S_k$.
        \end{IEEEproof}


    \subsection{Proof of Theorem~\ref{thm:choose_entries_bound}}
    \label{appen:proof_choose_entries_bound}
        Let $\cZ$ be the set of indices of the $s$-th smallest probabilities. 
        Consider a probability matrix $P' \in \R^{n \times m}$ that satisfies
        \begin{align*}
            P'_{ij} = 
            \begin{cases}
                \frac12  & (i,j) \in \cZ \\
                P_{ij} & (i,j) \notin \cZ.
            \end{cases}
        \end{align*}
        The matrix $P'$ agrees with $P$ on all entries in $\cZ^{\complement}$ and equals $\frac12$ otherwise.
        WLOG, assume $p_{(nm-s)} \le \frac12$. (Otherwise, change $\frac12$ to $p_{(nm-s)}$ in the above definition.) By this construction, we know $\min_{ij} P'_{ij} = p_{(nm-s)}$. 
        Let $E$ be the event that the following bound holds 
        \begin{align*}
                \norm{\hat A-A^{\ast}}_{\infty}  \lesssim \eta^{2} \kappa^{4} \left( \left\Vert A^{\ast}\right\Vert _{\infty} + \sigma\right) \sqrt{\frac{ (n+m) \log (1/\delta)}{p_{(nm-s)}}}.
            \end{align*}
        Invoking Theorem~\ref{thm:uniformization_bound}, we know that $$\mathbb{P}_{P'}(E^{\complement}) \le \delta.$$
        On the other hand, for each subset $\mathcal{S} \subseteq \mathcal{Z}$, let $F_{\mathcal{S}}$ denote the event that the entries in $\mathcal{S}$ are observed and those in $\mathcal{Z}\setminus\mathcal{S}$ are unobserved; note that $\mathbb{P}_{P'}(F_{\mathcal{S}}) = 2^{-s}$. By the law of total probability, we have 
        \begin{align*}
            \mathbb{P}_{P'}(E^{\complement}) 
            &= \sum_{\mathcal{S}\subseteq \mathcal{Z}} \mathbb{P}_{P'}(E^{\complement}\mid F_{\mathcal{S}}) \cdot \mathbb{P}_{P'}(F_{\mathcal{S}}) \\
            &= \sum_{\mathcal{S}\subseteq \mathcal{Z}} \mathbb{P}_{P}(E^{\complement}\mid F_{\mathcal{S}}) \cdot 2^{-s},
        \end{align*}
        where the last step follows from the fact that $\mathbb{P}_{P'}(\cdot \vert F_{\mathcal{S}}) = \mathbb{P}_{P}(\cdot \vert F_{\mathcal{S}}),\forall \mathcal{S}$ since $P$ and $P'$ are identical on entries outside $\mathcal{Z}$ and the observations are independent across entries. Combining the last two display equations, and lower bounding the sum by one summand, we have 
        \[
            \mathbb{P}_{P} (E^\complement \vert F_{\mathcal{S}}) \le 2^s \delta, \quad \forall \mathcal{S} \subseteq \mathcal{Z}
        \]
        and hence $ \mathbb{P}_{P} (E^\complement ) \le 2^s \delta$. Letting $\delta' = 2^s \delta$, we conclude that under $P$, with probability at least $1-\delta'$, it holds that 
        \begin{align*}
                \norm{\hat A-A^{\ast}}_{\infty}  \lesssim \eta^{2} \kappa^{4} \left( \left\Vert A^{\ast}\right\Vert _{\infty} + \sigma\right) \sqrt{\frac{ (n+m) \log (2^s/\delta')}{p_{(nm-s)}}}.
        \end{align*}

\section{Proofs for Section~\ref{sec:theoretical_guarantees}}

    \subsection{Proof of Theorem~\ref{thm:upper_bound}}
    \label{appen:proof_upper_bound}
    
        We first invoke Lemma~\ref{lem:low-rank_signal} and~Lemma~\ref{lem:bounded_signal} to obtain $M^\ast$ is both rank-$r$ and bounded, with probability at least $1-\delta/3$. 
        For entry $(i,j)$, we have a corresponding submatrix $W = M_{[k^\ast(i,j)] \cup \{i\}, [k^\ast(i,j)] \cup \{j\}}$ of size $k^\ast \times k^\ast$, $(k^\ast+1) \times k^\ast$, or $k^\ast \times (k^\ast+1)$. We apply Lemma~\ref{lem:incoherence_condition} to $W$ and get the incoherence and condition number guarantee with probability at least $1-\delta/3$:
        \begin{align*}
            \kappa(W) \le \frac{1 + C \sqrt \frac{r + \log(6(n+m+2)/\delta)}{k^\ast}}{1 - C \sqrt \frac{r + \log(6(n+m+2)/\delta)}{k^\ast}}.
        \end{align*}
        and
        {\scriptsize
        \begin{equation*}
            \begin{split}
                \norm{U^W}_{2 \to \infty} &\le C \left(  \sqrt{\frac{r}{k^\ast}}  + \sqrt{\frac{ \log(6(n+m+2)/\delta)}{k^\ast}} \right) \\
            \norm{V^W}_{2 \to \infty} &\le C \left( \sqrt{\frac{r}{k^\ast}}  + \sqrt{\frac{ \log(6(n+m+2)/\delta)}{k^\ast}} \right).
            \end{split}
        \end{equation*}
        }
        
        We consider three separate cases. 
        If $k^\ast \ge i$ and $k^\ast \ge j$, it means the smallest probability in $P_{[k^\ast] \cup \{i\}, [k^\ast] \cup \{j\}}$ is $P_{k^\ast, k^\ast}$, which is precisely $p^\ast(i,j)$. In this case, we invoke Theorem~\ref{thm:uniformization_bound} and the union bound to get
        \[
            \card{\hat M_{ij} - M^\ast_{ij}} \lesssim r \left( r + \sigma \right) \sqrt{\frac{\log ^5 (n/\delta) }{k^\ast(i,j) p^\ast(i,j)}},
        \]
        with probability at least $1-\delta$.
        If $k^\ast < i$ and $k^\ast < j$, then the smallest probability in $P_{[k^\ast] \cup \{i\}, [k^\ast] \cup \{j\}}$ becomes $P_{k^\ast k^\ast}$ and the second smallest is $\min \{ P_{i k^\ast}, P_{k^\ast j }\} = p^\ast(i,j)$. We apply Theorem~\ref{thm:choose_entries_bound} and the union bound to get
        \[
            \card{\hat M_{ij} - M^\ast_{ij}}  \lesssim r \left( r + \sigma \right) \sqrt{\frac{\log ^5 (n/\delta) }{k^\ast (i,j) p^\ast(i,j)}},
        \]
        with probability at least $1-\delta$.
        In the complementary case, we assume $i > k^\ast$ and $j \le k^\ast$ without loss of generality. The smallest probability in $P_{[k^\ast] \cup \{i\}, [k^\ast] \cup \{j\}}$ is  $P_{ik^\ast} = p^\ast (i,j)$. Apply Theorem~\ref{thm:uniformization_bound} and the same conclusion follows.



    \subsection{Proof of Theorem~\ref{thm:lower_bound}}
        \label{appen:proof_lower_bound}
        Recall the standard linear regression problems: $y = X \theta^\ast + \varepsilon$, with known covariates $X \in \R^{d \times r}$, unknown parameter $\theta^\ast \in \R^r$ and i.i.d.~Gaussian noise $\varepsilon_i \sim N(0,\sigma^2)$. 
        The minimax risk of estimating $\theta^\ast$ is
             \begin{equation}
                \label{eq:linear_regression_lower_bound}
                \inf_{\hat \theta} \sup_{\theta^\ast} \mathbb{E} \left[ \twonorm{\hat \theta - \theta^\ast}^2 \right] \ge \frac{C r \sigma^2 }{d}  \opnorm{\frac{1}{\sqrt{n}}X}.
            \end{equation}
        The proof can be found in~\cite{wainwright2019high-dim_stats}, which is an application of Fano's method. 
        
        Fix $i \in [n], j \in [m]$. Assume $\{a_k^\ast\}_{k\in[n]}$ is known and estimate $b_j^\ast$ using $\{ Y_{kj}, k \in [n]\}$.
        Applying~(\ref{eq:linear_regression_lower_bound}) yields:
        \begin{equation}
            \label{eq:matrix_completion_lower_bound}
                \inf_{\hat b_j} \sup_{b_j^\ast \in \R^r} \mathbb{E}[(\hat b_j - b_j^\ast )^2] \ge \frac{C r \sigma^2}{ \sum_{i'} P_{i'j}},
        \end{equation}
        with probability at least $\frac12$. The case where the column latent factors are known is symmetric. 
        We prove~(\ref{eq:matrix_completion_lower_bound}) and the desired result~(\ref{eq:lower_bound_ij}) follows.
        Let $X_k = \ind_{ \{ (k,j)\text{ is observed} \} }$ for all $k\in[n]$.
        We have $\mathbb{E}[X_k] = P_{kj}$. Applying Markov's inequality, we get
        \[
            \mathbb{P} \left(  \sum_k X_k \ge 2 \sum_{k} P_{kj} \right) \le \frac12.
        \]
        Let $\mathcal{K} = \{k \in [n] : X_k = 1\}$ be the set of observed entries.
        Applying (\ref{eq:linear_regression_lower_bound}) with $y = Y_{\mathcal{K} j}$, $X =(a_k^\ast)_{k \in \mathcal{K}}$, $\theta^\ast = b_j^\ast$ and $d = \sum_{k} X_k = \card{\mathcal{K}}$, we obtain
        \begin{equation*}
                \inf_{\hat b_j} \sup_{b_j^\ast} \mathbb{E}[(\hat b_j - b_j^\ast )^2] \ge \frac{C r \sigma^2}{\sum_k X_k} \ge \frac{C r \sigma^2}{2 \sum_k P_{kj}},
        \end{equation*}
        with probability at least $\frac12$ over the randomness of $\mathcal{K}$ and $\{a_k^\ast\}$.

\end{document}